\pdfoutput=1
\documentclass[11pt]{article}

\usepackage[preprint]{acl}

\usepackage{times}
\usepackage{latexsym}
\usepackage{tabularx}
\usepackage{multirow}
\usepackage{booktabs}
\usepackage{arydshln}
\usepackage{tikz}
\usetikzlibrary{positioning}
\usepackage{color}
\usepackage{amssymb}
\usepackage{pgfplots}
\usepackage{CJKutf8}

\newcommand{\tokensep}{\textcolor{red}{$\blacklozenge$}}

\usepackage[T1]{fontenc}
\usepackage[utf8]{inputenc}

\usepackage{microtype}

\usepackage{inconsolata}

\usepackage{graphicx}
\usepackage{here}

\title{Step-by-step Instructions and a Simple Tabular Output Format Improve the Dependency Parsing Accuracy of LLMs}

\author{~~~~~~~~~~~~~~~~~~ Hiroshi Matsuda ~~~~~~~~~~~~ Chunpeng Ma \\
  ~~~~~~~~~~~~~~~~~~ Megagon Labs, Tokyo, \\
  ~~~~~~~~~~~~~~~~~~ Recruit Co., Ltd. \\
  ~~~~~~~~~~~~~~~~~~ \texttt{\{hiroshi\_matsuda,~ma.chunpeng\}@megagon.ai} \\\And
  ~~~~~~~~~~~~~~~~~~ Masayuki Asahara \\
  ~~~~~~~~~~~~~~~~~~ National Institute for Japanese \\
  ~~~~~~~~~~~~~~~~~~ Language and Linguistics \\
  ~~~~~~~~~~~~~~~~~~ \texttt{masayu-a@ninjal.ac.jp} \\}

\begin{document}

\maketitle

\begin{abstract}
Recent advances in large language models (LLMs) have enabled impressive performance in various tasks. However, standard prompting often struggles to produce structurally valid and accurate outputs, especially in dependency parsing. We propose a novel step-by-step instruction strategy, where universal part-of-speech tagging precedes the prediction of syntactic heads and dependency labels, and a simplified CoNLL-U like output format, our method achieves state-of-the-art accuracy on Universal Dependencies datasets across 17 languages without hallucination or contamination. We further show that multilingual fine-tuning simultaneously improves cross-language generalization performance. Our results highlight the effectiveness of explicit reasoning steps in LLM-based parsing and offer a scalable, format-consistent alternative to bracket-based approaches.
\end{abstract}

\section{Introduction}

\begin{figure}[t]
    \centering
    \includegraphics[width=\linewidth]{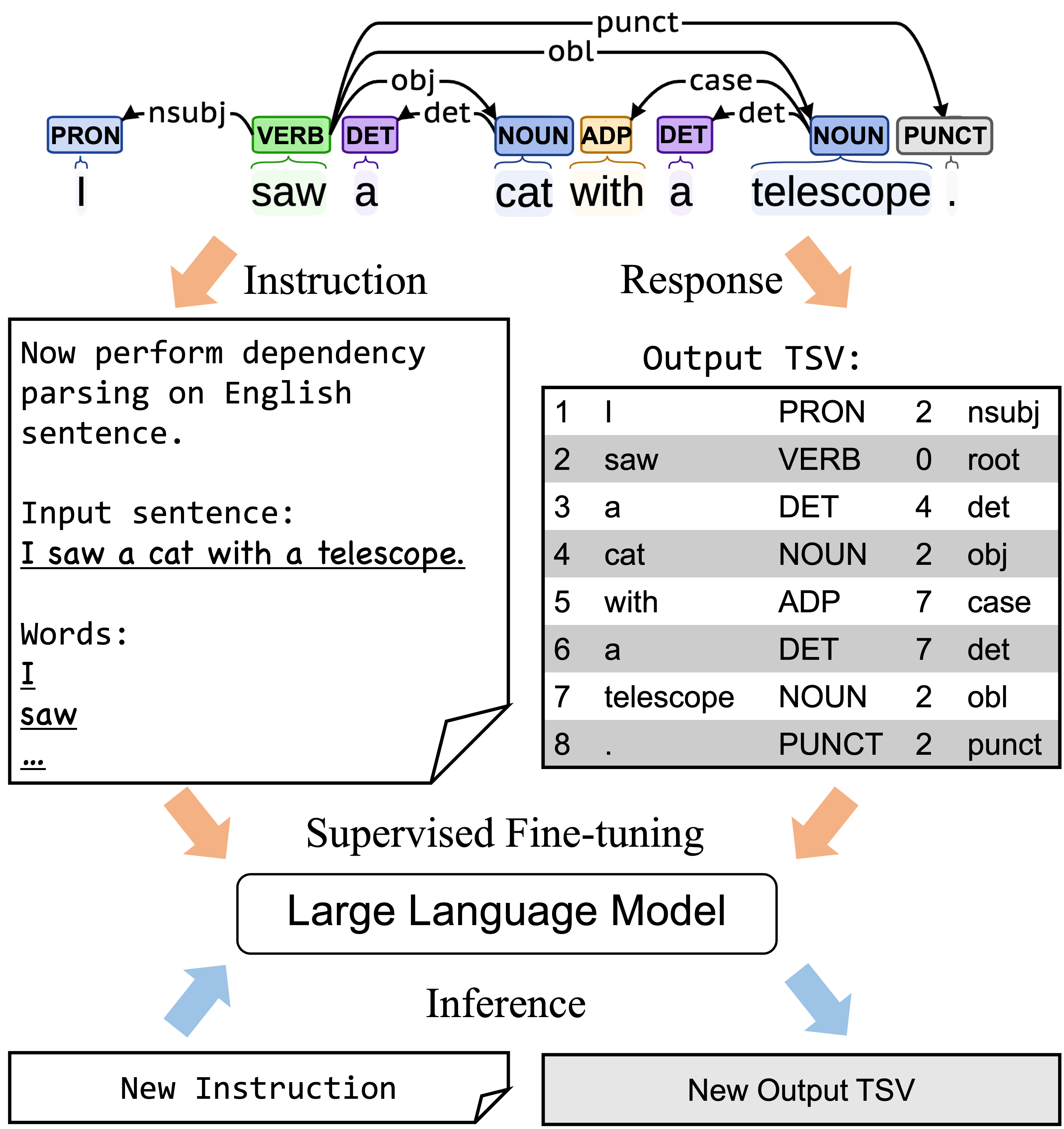}
    \caption{Framework of the proposed method.}
    \label{fig:framework}
\end{figure}

Recent advances in large language models (LLMs) have dramatically reshaped the landscape of natural language processing; however, their potential for syntactic analysis -- particularly dependency parsing -- remains underexplored.
Furthermore, it is desirable to systematically investigate prompting and fine-tuning techniques that enhance the performance of LLM-based dependency parsing.

In this work, we examine how fine-tuned LLMs can be effectively guided to perform accurate dependency parsing using simple, structured instruction prompts.
Specifically, we design a single-turn supervised fine-tuning setup where the input sentence is accompanied by a tabular output format based on a minimal subset of the \texttt{CoNLL-U}\footnote{\href{https://universaldependencies.org/format.html}{https://universaldependencies.org/format.html}}, which is the standard format of Universal Dependencies (UD) treebanks \citep{nivre-2020} as in Figure \ref{fig:framework}.
This table-based representation not only improves format validity and readability, but also facilitates learning non-projective structures.

The results of our preliminary experiments using \texttt{UD\_English-EWT}\footnote{\href{https://universaldependencies.org/treebanks/en_ewt/}{https://universaldependencies.org/treebanks/en\_ewt/}} are summarized in Table~\ref{tab:eval-cot-steps}.
First, we found that performing SFT with a single-step prompt yielded accuracy comparable to or better than that of \texttt{UDPipe 2.0} \citep{straka-2018}. 
Next, we introduced a step-by-step prompting strategy in a Chain-of-Thought style \citep{wei-2022}.
Specifically, we first predict UPOS tags, then syntactic heads and dependency relations.
We observed that step-by-step prompts leads to substantial gains in both unlabeled attachment score (UAS) and labeld attachment socre (LAS).

Despite using a very simple prompt, we observed fairly high parsing accuracy, prompting us to investigate the possibility of data contamination (refer Appendix~\ref{app:contamination-verification} for details).
Based on our analysis, we found no evidence of contamination in the prediction of syntactic heads, and dependency relations by the models used in this study for the test set of \texttt{UD\_English-EWT r2.15}.
However, we suspect that the part-of-speech tagging may have been exposed to the models during its pre- and mid-training\footnote{\href{https://vintagedata.org/blog/posts/what-is-mid-training}{https://vintagedata.org/blog/posts/what-is-mid-training}} phases.

\begin{table}[t]
\centering
\resizebox{\linewidth}{!}{
\begin{tabular}{l|cccc}
\hline
 & Token Recall & UPOS & UAS & LAS\\
\hline
\texttt{UDPipe 2.0} & 100.0 & 97.5 & 93.4 & 91.5 \\
\hline
\texttt{gpt-4o-mini-2024-07-18}: \\
Chain-of-Thought Steps \\
1 (UPOS+HEAD+DEPREL)&	98.5&	98.4&	93.3&	91.4 \\
2 (UPOS; HEAD+DEPREL)&	99.1&	\textbf{98.6}&	94.8&	93.1 \\
3 (UPOS; HEAD; DEPREL)&	\textbf{99.6}&	\textbf{98.6}&	\textbf{95.1}&	\textbf{93.5} \\
\hline
\end{tabular}
}
\caption{Preliminary experiment on evaluating Chain-of-Thought effect in \texttt{UD\_English-EWT r2.15}. We performed all steps within a single-turn prompt. The example prompts are presented in Appendix \ref{app:prompt_examples}.}
\label{tab:eval-cot-steps}
\end{table}

\section{Related Work}

Linearization techniques are essential for both constituency parsing \citep{vinyals-2015, ma-2017} and dependency parsing \citep{li-2018, hromei-2024} using sequence-to-sequence model with bracket-based representations, illustrated in Table \ref{tab:previous-research}.

In generative parsing using bracket-based representations, the tree structure in the output text is often invalid, which is one of the factors that reduces the accuracy of parsing, resulting in additional recovery procedure \citep{bai-2023}, or even redesign the topology of neural networks to ensure the output validity \citep{dyer-2015,gomez-2018}.

\begin{table}
    \centering
    \footnotesize
    \resizebox{1.0\linewidth}{!}{
    \begin{tabularx}{\linewidth}{X|X}
    \hline
    \citet{vinyals-2015} & \citet{hromei-2024}\\
    \hline
    \includegraphics[width=\linewidth]{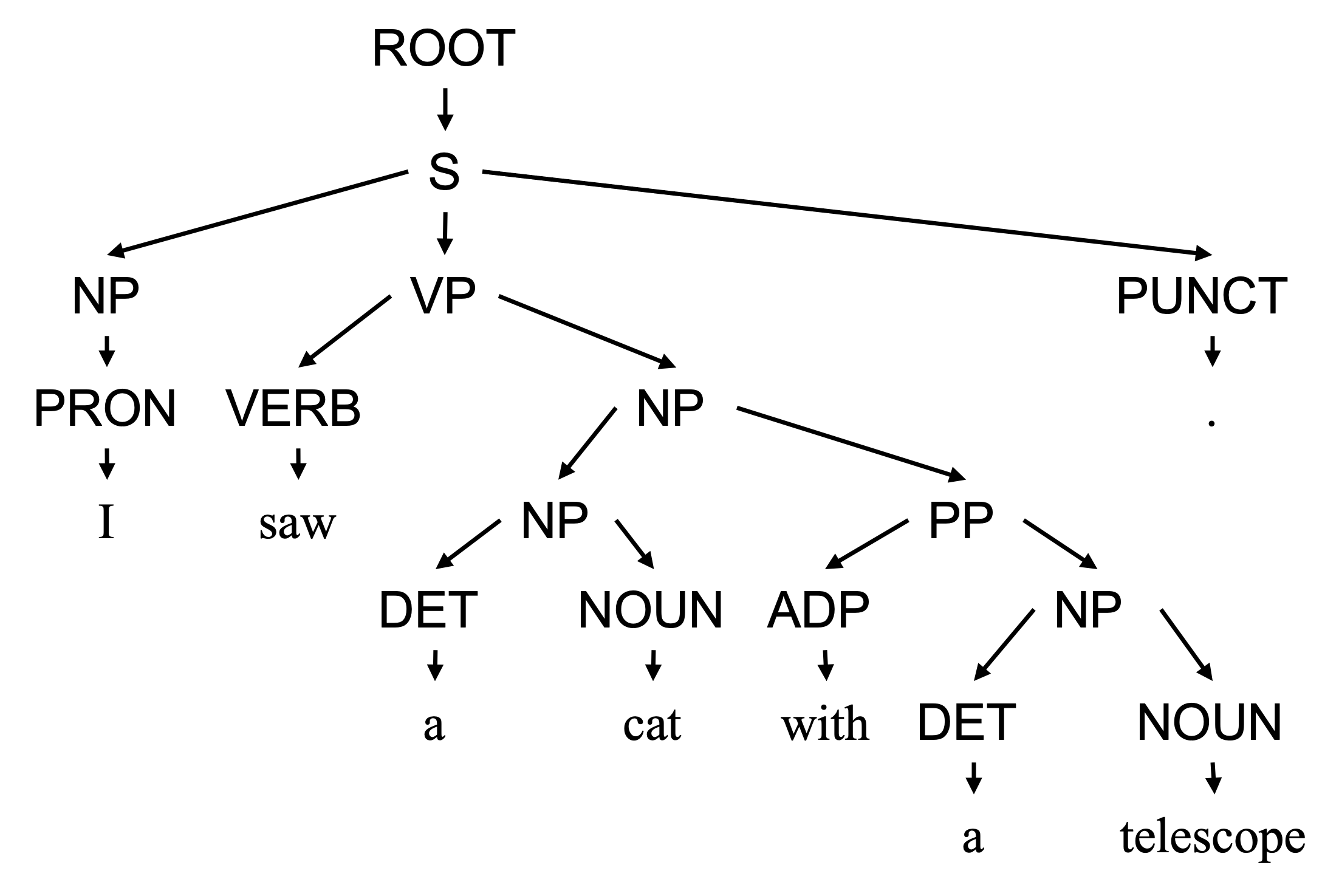} & \includegraphics[width=\linewidth]{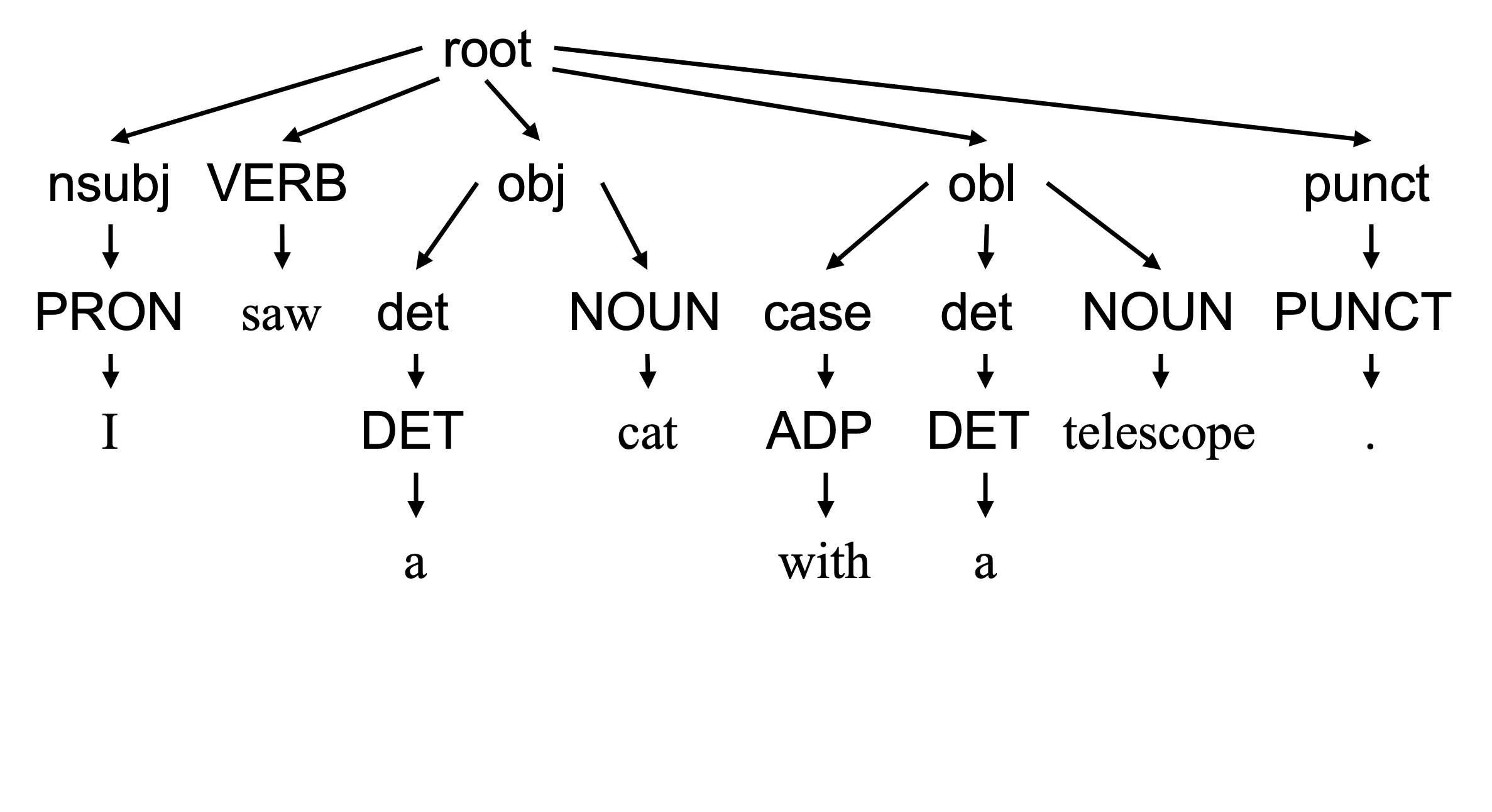}\\
    (ROOT\tokensep(S\tokensep & [root\tokensep[nsubj\tokensep[PRON\tokensep[I\tokensep]]]\\
    \,(NP\tokensep PRON\tokensep)$_{NP}$\tokensep & \,\tokensep[VERB\tokensep[saw\tokensep]]\tokensep\\
    \,(VP\tokensep VERB\tokensep(NP\tokensep & \,[obj\tokensep[det\tokensep[DET\tokensep[a\tokensep]]]\tokensep\\
    \;(NP\tokensep DET\tokensep NOUN\tokensep)$_{NP}$\tokensep & \;[NOUN\tokensep[cat\tokensep]]]\tokensep\\
    \;(PP\tokensep ADP\tokensep & \,[obl\tokensep[case\tokensep[ADP\tokensep[with\tokensep]]]\\
    \,\,(NP\tokensep DET\tokensep NOUN\tokensep)$_{NP}$\tokensep & \;\tokensep[det\tokensep[DET\tokensep[a\tokensep]]]\tokensep\\
    \;)$_{PP}$\tokensep)$_{NP}$\tokensep)$_{VP}$\tokensep & \;[NOUN\tokensep[telescope\tokensep]]]\tokensep \\
    \,PUNCT\tokensep)$_S$\tokensep)$_{ROOT}$ & \,[punct\tokensep[PUNCT\tokensep[.\tokensep]]]] \\
    \hline
    \end{tabularx}
    }
    \caption{Comparison of bracket-based linearization methods. Syntactic elements are separated by ``\tokensep''.}
    \label{tab:previous-research}
\end{table}

\section{Approach}

In this section, we describe a table-based representation of dependency structures, similar to the \texttt{CoNLL-U} format, and explain how to construct instruction prompts for dependency parsing.

\subsection{Table-based representation}

Recent large language models (LLMs) have significantly improved their ability to output in structured formats such as JSON or CSV, enabling function calling for flexible interaction with external services\footnote{\href{https://platform.openai.com/docs/guides/function-calling}{https://platform.openai.com/docs/guides/function-calling}}.  
This capability facilitates the direct handling of tabular structures such as \texttt{CoNLL-U}, potentially allowing LLMs to generate parse results with higher structural validity compared to the bracket-based representations employed in prior studies.  

In this work, we adopt a table-based representation that extracts only the essential fields -- ID, FORM, UPOS, HEAD, and DEPREL -- from the \texttt{CoNLL-U} format, as illustrated in the output TSV in Figure~\ref{fig:framework}.  
A further advantage of the table-based approach is its ability to naturally represent non-projective dependency structures using index-based head references.
However, it should be noted that table-based representations can represent circular references and multiple roots.
As we demonstrate in the next section, the tabular outputs generated by the LLMs were mostly well-formed, and the validity errors were fairly rare on the \texttt{UD\_English-EWT} r2.15 test set.
Furthermore, the table-based representation offers an advantage in recovery processing, as it can accurately recover word indices and forms as long as the number of records and the field structure are correctly output.

\subsection{Step-by-step instruction prompts}

We began our preliminary experiments using the simple single-step prompt illustrated in Figure~\ref{fig:framework}.  
Through iterative refinement, we found that parsing the UPOS tags first, followed by the HEAD and DEPREL fields in a step-by-step manner, led to improved accuracy.  
Accordingly, the experiments presented in next chapter employ a three-step Chain-of-Thought prompting strategy, processing the elements in the order of UPOS, HEAD, and DEPREL. Representative examples of these prompt templates are provided in the Appendix~\ref{app:prompt_examples}.

\begin{table*}[h]
\centering
\small
\begin{tabular}{l|c|cccc|cccc}
\hline
\multirow{3}{*}{Models} & \# of Parameters & \multicolumn{4}{c|}{\texttt{UD\_English-EWT r2.2}} & \multicolumn{4}{c}{\texttt{UD\_English-EWT r2.15}} \\
& Trainable vs. & Token & \multirow{2}{*}{UPOS} & \multirow{2}{*}{UAS} & \multirow{2}{*}{LAS} & Token & \multirow{2}{*}{UPOS} & \multirow{2}{*}{UAS} & \multirow{2}{*}{LAS} \\
& Transformer & Recall &  &  &  & Recall &  &  & \\
\hline
Baselines: & & & & & & & & & \\
\texttt{UDPipe 2.0}               & 30.1M$^\dag$ / 168M & -     & -     & -      & -      & - & 97.5 & 93.4   & 91.5 \\
\texttt{Hexatagger bert-base}     & 222M  / 178M & - & -     & 91.4   & 88.7   & - & -    & 93.5   & 91.4 \\
\quad\texttt{(+ Gold POS)}        & 224M  / 178M & - & -     & (93.3) & (91.1) & - & -    & (93.8) & (92.1) \\
\texttt{U-DepP Llama-2-13b-hf}$^*$& 31.6M / 13.0B & 96.3 & -    & 88.9   & 86.6   & 95.3 & -    & 92.3   & 90.1 \\
\texttt{U-DepP Llama-3.1-8B}$^*$  & 22.0M / 8.05B & 98.4 & -    & 90.7   & 88.4   & 97.8 & -    & 92.0   & 90.2 \\
\hline
Our methods: & & & & & & & & \\
\texttt{gpt-4o-mini-2024-07-18}   & (undisclosed) & 99.6  & 97.7 & 93.2 & 91.1 & 99.5  & 98.3 & 94.9 & 93.3 \\
\texttt{gpt-4o-2024-08-06}        & (undisclosed) & 99.7  & \textbf{98.1} & 93.7 & 91.7 & \textbf{100.0} & 98.5 & 95.2 & 93.5 \\
\texttt{gemma-2-2b}$^*$           & 10.4M / 2.62B & 99.7 & 98.0 & 93.2 & 91.1 & 99.9 & 98.3 & 94.4 & 92.7 \\
\texttt{Qwen2.5-7B}$^*$           & 20.2M / 7.64B & 99.5  & 97.9 & 93.1 & 91.0 & 99.5  & 98.3 & 94.7 & 93.0 \\
\texttt{Llama-3.1-8B}$^*$         & 21.0M / 8.05B & 99.4  & 97.8 & 93.4 & 91.3 & \textbf{100.0} & 98.4 & 94.8 & 93.1 \\
\texttt{gemma-2-9b}$^*$           & 27.0M / 9.27B & \textbf{99.8}  & \textbf{98.1} & \textbf{93.8} & \textbf{91.9} & \textbf{100.0} & \textbf{98.6} & \textbf{95.5} & \textbf{94.1} \\
\hline
\end{tabular}
\caption{Evaluation of various models in \texttt{UD\_English-EWT r2.2} and \texttt{r2.15}. Best scores are highlighted in \textbf{bold}. The scores for \texttt{UDPipe 2.0} are taken from its official documentation. The scores for \texttt{Hexatagger} and \texttt{U-DepPLLaMA} are the results of our reproduction experiments. The scores in the row \texttt{(+ Gold POS)} are provided for reference, as they use gold POS tags. The LoRA-SFT models are marked by “$^*$”. “$^\dag$” indicates that the value is estimated from the size of distributed model archive.}
\label{tab:eval-detailed-ewt}
\end{table*}

{
\tabcolsep = 2pt
\begin{table*}[h!]
\centering
\resizebox{\linewidth}{!}{
\begin{tabular}{l|ccc:cc|ccc:ccc:ccc:ccc||ccc}
\hline
& \multicolumn{5}{c|}{Baselines} & \multicolumn{12}{c||}{Ours: Monolingual} & \multicolumn{3}{c}{Ours: Multilingual} \\
\hline
\multirow{2}{*}{Dataset}& \multicolumn{3}{c:}{\texttt{UDPipe 2.0}} & \multicolumn{2}{c|}{\texttt{Hexatagger(+Gold POS)}} & \multicolumn{3}{c:}{\texttt{gemma-2-2b}$^*$} & \multicolumn{3}{c:}{\texttt{Qwen2.5-7B}$^*$} & \multicolumn{3}{c:}{\texttt{Llama-3.1-8B}$^*$} & \multicolumn{3}{c||}{\texttt{gemma-2-9b}$^*$} & \multicolumn{3}{c}{\texttt{gemma-2-9b}$^*$} \\
&    UPOS& UAS&  LAS~&  UAS&        LAS&        UPOS& UAS&  LAS&  UPOS& UAS&  LAS&  UPOS& UAS&  LAS&  UPOS& UAS&  LAS&  UPOS& UAS&  LAS \\
\hline
ar\_padt&      \underline{97.0}& 88.1& 83.7~~&~~86.8 (87.9)& 81.8 (83.9)& 95.4 &87.7 & 83.1 & 96.1& 88.7& 84.3& \textbf{97.1}& \textbf{90.0}& \textbf{85.5}& 96.2& \underline{89.8}& \underline{85.4}& 96.2& 89.9& \textbf{85.7}\\
bg\_btb&       \underline{99.3}& 95.3& 92.6~~&~~94.7 (95.4)& 92.0 (92.4)& \underline{99.3} & 95.3 & 92.9 & 99.2& 95.2& 92.7& 99.1& \underline{95.7}& \underline{93.2}& \textbf{99.5}& \textbf{96.9}& \textbf{94.7}& \textbf{99.5}& \textbf{97.0}& \textbf{94.8}\\
ca\_ancora&    \underline{99.2}& 94.9& 93.4~~&~~94.8 (95.1)& 93.1 (93.7)& 98.9 & 94.5 & 92.9 & 99.0& 94.7& 93.1& 99.1& \underline{95.1}& \underline{93.6}& \textbf{99.3}& \textbf{95.7}& \textbf{94.5}& \textbf{99.3}& \textbf{95.7}& 94.3\\
cs\_pdt$^\Diamond$&\underline{99.4}& \underline{95.7}& \underline{94.3}~~&~~94.3 (94.7)& 92.5 (92.9)& 99.3 & 95.1 & 93.6 & 99.3& 95.2& 93.8& 99.3& 95.5& 94.2& \underline{99.4}& \underline{95.7}& \textbf{94.4}& 99.3& 94.9& 93.3\\
de\_gsd&       \underline{97.1}& 89.2& 85.5~~&~~87.9 (90.2)& 83.8 (86.1)& 97.0 & 89.1 & 85.3 & \underline{97.1}& \underline{89.7}& 85.9& \underline{97.1}& \underline{89.7}& \underline{86.1}& \textbf{97.4}& \textbf{90.2}& \textbf{86.5}& \textbf{97.4}& \textbf{90.3}& \textbf{86.7}\\
en\_ewt&       97.5& 93.4& 91.5~~&~~93.5 (93.8)& 91.4 (92.1)& 98.3 & 94.4 & 92.7 & 98.3& 94.7& 93.0& \underline{98.4}& \underline{94.8}& \underline{93.1}& \textbf{98.6}& \textbf{95.5}& \textbf{94.1}& \textbf{98.6}& \textbf{95.5}& \textbf{94.1}\\
es\_ancora&    \underline{99.1}& 94.0& 92.4~~&~~93.4 (94.6)& 91.6 (92.9)& \underline{99.1} & 93.7 & 92.0 & \underline{99.1}& 94.1& 92.6& \underline{99.1}& \underline{94.2}& \underline{92.7}& \underline{99.1}& \textbf{94.6}& \textbf{93.2}& \textbf{99.2}& \textbf{95.1}& \textbf{93.7}\\
fr\_gsd&       98.5& 95.0& 93.2~~&~~94.4 (96.0)& 93.3 (94.6)& 98.6 & 95.6 & 94.0 & \underline{98.7}& \underline{95.9}& \underline{94.4}& 98.5& 95.6& 93.9& \underline{98.7}& \textbf{96.7}& \textbf{95.3}& \textbf{98.8}& \textbf{96.7}& 95.2\\
it\_isdt&      \textbf{98.8}& 95.1& 93.4~~&~~95.7 (96.5)& 94.0 (95.2)& 98.6 & 95.6 & 94.0 & 98.4& 94.8& 93.2& \underline{98.7}& \underline{95.8}& \underline{94.4}& \underline{98.7}& \textbf{96.0}& \textbf{94.6}& \textbf{98.8}& \textbf{96.4}& \textbf{94.9}\\
ja\_gsd&       98.6& \underline{95.1}& \underline{94.2}~~&~~94.6 (95.2)& 93.3 (94.6)& 98.7 & 94.4 & 93.1 & \underline{98.8}& 94.8& 93.6& 98.5& 94.8& 93.6& \textbf{99.0}& \textbf{95.7}& \textbf{94.9}& 98.9& 95.5& 94.5\\
ko\_gsd&       96.7& 88.9& 85.2~~&~~87.4 (87.9)& 83.6 (85.1)& 96.5 & 87.4 & 83.9 & 96.8& \underline{89.3}& \underline{86.5}& \underline{96.9}& 88.8& 85.9& \textbf{97.2}& \textbf{90.3}& \textbf{87.3}& \textbf{97.2}& \textbf{90.6}& \textbf{87.6}\\
nl\_alpino&    \underline{98.2}& \underline{94.9}& \underline{92.9}~~&~~93.3 (94.1)& 90.6 (91.5)& 98.1 & 94.0 & 91.3 & 97.4& 94.0& 91.6& 97.3& 93.9& 91.6& \textbf{98.4}& \textbf{95.7}& \textbf{93.9}& \textbf{98.5}& 95.6& 93.7\\
no\_bokmaal&   98.6& 94.7& 93.4~~&~~\textbf{95.9} (96.5)& \underline{94.3} (94.8)& \underline{98.7} & 94.7 & 93.5 & 98.5& 94.2& 92.9& \underline{98.7}& 95.0& 93.8& \textbf{98.8}& \underline{95.7}& \textbf{94.8}& 98.7& 95.5& 94.6\\
ro\_rrt&       \underline{98.1}& 92.6& 89.1~~&~~91.9 (92.9)& 87.9 (89.0)& \underline{98.1} & 92.5 & 88.9 & 97.9& 92.7& 89.2& 97.7& \underline{93.1}& \underline{89.7}& \underline{98.1}& \textbf{94.4}& \textbf{91.3}& \textbf{98.4}& 94.3& \textbf{91.3}\\
ru\_syntagrus& 98.8& 94.5& 92.3~~&~~94.1 (95.1)& 91.7 (92.7)& 98.8 & 95.2 & 93.1 & \underline{98.9}& 95.3& 93.4& \underline{98.9}& \underline{95.5}& \underline{93.5}& \textbf{99.0}& \textbf{95.8}& \textbf{93.8}& 98.6& 95.0& 92.7\\
sl\_ssj&       98.8& 94.5& 92.9~~&~~94.0 (94.8)& 92.3 (93.2)& 98.5 & 94.0 & 92.4 & 98.7& 94.1& 92.4& \underline{98.9}& \underline{94.7}& \underline{93.2}& \textbf{99.0}& \textbf{95.9}& \textbf{94.6}& \textbf{99.0}& \textbf{95.9}& \textbf{94.7}\\
zh\_gsdsimp&   95.8& 86.7& 83.6~~&~~87.4 (90.1)& 84.3 (87.9)& 96.4 & 86.5 & 83.6 & 96.5& 87.2& 84.4& \underline{96.8}& \underline{87.7}& \underline{84.7}& \textbf{97.5}& \textbf{89.3}& \textbf{86.9}& 97.3& \textbf{89.7}& \textbf{87.3}\\
\hline
\end{tabular}
}
\caption{Evaluation results on various UD r2.15 datasets. For each language, best scores among the baselines and our monolingual models are shown in \textbf{bold}, with ties and second-best scores \underline{underlined}. Additionally, scores from our multilingual model that outperform the baselines and monolingual models are also highlighted. The scores for \texttt{UDPipe 2.0} are taken from its official documentation. The scores for \texttt{Hexatagger} are the results of our reproduction experiments. The scores in the brackets are provided for reference, as they use gold POS tags. LoRA-SFT models are marked by “$^*$”. “$^\Diamond$” indicates use of a language-specific pre-trained model in \texttt{UDPipe 2.0}.}
\label{tab:eval-17-lang}
\end{table*}
}

\section{Experiments}

We conducted both supervised fine-tuning (SFT) with Low-Rank Adaptation (LoRA) \citep{hu-2022} and inference experiments for open LLMs on a high-performance cloud service\footnote{
Experiments were conducted on a Google Cloud A2 Ultra instance with 8 $\times$ NVIDIA A100 GPUs (80GB each), 96-core Intel Xeon CPUs @ 2.20GHz, 1,360GB RAM, and 5TB of SSD storage.
The software environment included: Ubuntu 22.04, CUDA 12.1, Python 3.11.9, PyTorch 2.5.1, Transformers 4.49.0, TRL 0.15.2, PEFT 0.14.0, OpenAI 1.68.2, Unsloth 2025.3.18, and vLLM 0.7.2.
}$^,$\footnote{The implementation used in the experiments is available on GitHub. \href{https://github.com/megagonlabs/llmpp}{https://github.com/megagonlabs/llmpp}}.
For OpenAI models, SFT was performed via the official web console\footnote{\href{https://platform.openai.com/docs/guides/fine-tuning}{https://platform.openai.com/docs/guides/fine-tuning}. The cost of fine-tuning the en\_ewt-r2.15 train set for 2 epochs was about \$52 for gpt-4o-mini and about \$430 for gpt-4o.}.
We explored SFT hyper-parameters\footnote{
\textbf{Open LLMs:} \texttt{num\_epochs}=3, \texttt{max\_seq\_length}=8192, \texttt{lr}=3e-4, \texttt{lr\_scheduler}=cosine\_with\_min\_lr, \texttt{min\_lr}=0.1, LoRA: \texttt{r}=8, \texttt{dropout}=0.05, \texttt{target\_modules}="all-linear" (embedding layers excluded). \textbf{OpenAI:} \texttt{num\_epochs}=2, \texttt{max\_seq\_length}=8192, \texttt{lr}=default.
} on the \texttt{UD\_English-EWT r2.15} development set and applied them to all experiments.
We used simple TSV recovery process only restores the ID and FORM on a row-by-row basis.

\subsection{Dataset}

We mainly used Universal Dependencies treebanks r2.15. For \texttt{UD\_English-EWT} (\texttt{en\_ewt}), we also used r2.2 for comparison with baseline methods.

\paragraph{For monolingual SFT.}

We used datasets for the following 17 languages to evaluate the parsing accuracy for each language:
\texttt{ar\_padt},
\texttt{bg\_btb},
\texttt{ca\_ancora},
\texttt{cs\_pdt},
\texttt{de\_gsd},
\texttt{en\_ewt},
\texttt{es\_ancora},
\texttt{fr\_gsd},
\texttt{it\_isdt},
\texttt{ja\_gsd},
\texttt{ko\_gsd},
\texttt{nl\_alpino},
\texttt{no\_bokmaal},
\texttt{ro\_rrt},
\texttt{ru\_syntagrus},
\texttt{sl\_ssj},
and \texttt{zh\_gsdsimp}.
Statistics for each dataset are provided in the Appendix~\ref{app:dataset-statistics}.

\paragraph{For multilingual SFT.}

To train a multilingual parsing model, we constructed a new dataset by gathering training sets from the datasets for the 17 languages above.
To reduce training time and costs, we downsampled \texttt{cs\_pdt} and \texttt{ru\_syntagrus} by 17\% to balance them with other language datasets.
The final training data consisted of 182,255 sentences and 3,889,494 tokens, which was used to train a multilingual model (denoted as \texttt{17\_multi} below).
Additionally, we evaluated the following 10 language datasets not included in the multilingual training data:
\texttt{el\_gdt},
\texttt{he\_htb},
\texttt{hi\_hdtb},
\texttt{hu\_szeged},
\texttt{id\_gsd},
\texttt{pt\_gsd},
\texttt{sv\_talbanken},
\texttt{tr\_imst},
\texttt{vi\_vtb},
and \texttt{zh\_gsd}.

\subsection{Baseline methods}

We compared our method against three strong baselines:
\texttt{UDPipe 2.0}\footnote{\href{https://ufal.mff.cuni.cz/udpipe/2/models}{https://ufal.mff.cuni.cz/udpipe/2/models}} \citep{straka-2018},  
\texttt{Hexatagger}\footnote{\href{https://github.com/rycolab/parsing-as-tagging}{https://github.com/rycolab/parsing-as-tagging}} \citep{amini-2023}, and  
\texttt{U-DepPLLaMA}\footnote{\href{https://github.com/crux82/u-deppllama}{https://github.com/crux82/u-deppllama}} \citep{hromei-2024}.
The reported scores for \texttt{UDPipe 2.0} were taken from its official documentation, while the results for \texttt{Hexatagger} and \texttt{U-DepPLLaMA} were reproduced in our environment using their publicly available implementations\footnote{The publicly available implementation of U-DepPLLaMA uses the precision as the accuracy, but we followed the UD convention and used F1-measure as the accuracy.}.
For Hexatagger, we report the accuracy under the setting that does not use gold POS tags (the accuracy when using gold POS tags is also provided as a reference).

\subsection{Evaluation of various models}

We conducted a comparative evaluation of baselines and our SFT method with various LLMs including
\texttt{gpt-4o-mini}\footnote{\href{https://platform.openai.com/docs/models/gpt-4o-mini}{https://platform.openai.com/docs/models/gpt-4o-mini}},  
\texttt{gpt-4o}\footnote{\href{https://platform.openai.com/docs/models/gpt-4o}{https://platform.openai.com/docs/models/gpt-4o}},  
\texttt{gemma-2-2b}\footnote{\href{https://huggingface.co/google/gemma-2-2b}{https://huggingface.co/google/gemma-2-2b}},
\texttt{Qwen2.5-7B}\footnote{\href{https://huggingface.co/Qwen/Qwen2.5-7B}{https://huggingface.co/Qwen/Qwen2.5-7B}},  
\texttt{Llama-3.1-8B}\footnote{\href{https://huggingface.co/meta-llama/Llama-3.1-8B}{https://huggingface.co/meta-llama/Llama-3.1-8B}}, and  
\texttt{gemma-2-9b}\footnote{\href{https://huggingface.co/google/gemma-2-9b}{https://huggingface.co/google/gemma-2-9b}}
 using \texttt{UD\_English-EWT}.

Results are summarized in Table~\ref{tab:eval-detailed-ewt}.
Overall, \texttt{gemma-2-9b} achieved the highest performance, followed closely by \texttt{gpt-4o}.
Beyond Table 3, circular references were rare, with only 3 cases found in the output of \texttt{Qwen2.5-7B}, and no multiple roots found in the output of either model on the test set.
These results highlight the favorable cost-performance trade-off of open LLMs, leading us to exclude OpenAI models from the subsequent experiments.

From the perspective of model parameter size, the pre-trained LLMs used in this experiment contain 2.6 to 9.3 billion parameters, which is several tens of times larger than the bert-base models used in the baselines.
However, the numbers of trainable LoRA parameters are relatively small, ranging from 10 to 27 million.
This suggests that LoRA-based SFT effectively leverages the capabilities of large, fixed-weight networks for dependency parsing tasks.
Moreover, the parsing accuracy appears to depend on the number of pre-training parameters, given a certain number of trainable parameters.

\subsection{Evaluation in 17 languages}

\paragraph{Monolingual SFT.}
We evaluated the proposed method in 17 UD languages to assess its monolingual performance.
Table~\ref{tab:eval-17-lang} shows the detailed results for each language.

The proposed method achieved the highest LAS in all 17 languages, and the highest UAS in 16, except Norwegian, indicating its overall effectiveness.
Among the open LLMs, \texttt{gemma-2-9b} demonstrated consistently strong performance, ranking first in 16 languages with the sole exception of Arabic.  
Due to lower tokenization efficiency in \texttt{ar\_padt} compared to other languages, the LLMs occasionally failed to output the complete analysis results within the available context length, particularly for long sentences.
However, the \texttt{Llama-3.1} tokenizer was approximately 20\% more efficient at tokenizing Arabic text than the \texttt{gemma-2} and \texttt{Qwen2.5} tokenizers, which contributing to higher accuracy.
This indicates a trade-off between efficiency and accuracy: as the number of Chain-of-Thought steps increases, the allowable input sentence length becomes more constrained by the maximum context length of the LLMs.

\begin{table}[h!]
\centering
\small
\begin{tabular}{lccc}
\hline
\multirow{2}{*}{Dataset}& \multicolumn{3}{c}{\texttt{gemma-2-9b 17-multi}} \\
& UPOS & UAS & LAS \\
\hline
el\_gdt	&93.1	&92.4	&86.3\\
he\_htb	&90.1	&83.3	&70.0\\
hi\_hdtb	&76.9	&72.9	&55.2\\
hu\_szeged	&87.1	&85.7	&75.1\\
id\_gsd	&86.5	&82.4	&66.6\\
pt\_gsd	&92.2	&86.4	&78.0\\
sv\_talbanken	&92.9	&89.6	&82.1\\
tr\_imst	&79.2	&67.9	&53.5\\
vi\_vtb	&81.8	&72.3	&57.7\\
zh\_gsd	&97.2	&89.5	&87.2\\
\hline
\end{tabular}
\caption{Evaluation results of our multilingual model on UD r2.15 datasets not used for training.}
\label{tab:eval-10-lang}
\end{table}

\paragraph{Multilingual SFT.}
An additional advantage of the proposed method is its compatibility with multilingual training.
The \texttt{gemma-2-9b 17-multi} model achieved comparable or higher accuracy than its monolingual counterparts, except in Czech and Russian, likely due to the down-sampling.

Table~\ref{tab:eval-10-lang} shows the evaluation results on 10 languages not included in the training data for \texttt{17\_multi}.
Among these, Greek and Swedish exhibited relatively high performance, indicating successful generalization from typologically or linguistically related languages.
This highlights the model’s ability to generalize across languages, a key strength of our method.

\subsection{Analysis}

\paragraph{Error analysis.}

We conducted an error analysis on Simplified Chinese, which showed the lowest UAS in monolingual evaluation.
Errors were primarily concentrated in nouns (27.8\%), verbs (24.8\%), and punctuation marks (16.1\%) for \texttt{gemma-2-9b}.
Most of these errors occurred in sentences containing multiple independent clauses—a structure more frequent in Chinese than in many other languages.  
Due to the structural parallelism among these clauses, an output that differs from the gold annotation is not necessarily incorrect.  

\begin{figure}[h!]
    \centering
    \includegraphics[width=1\linewidth]{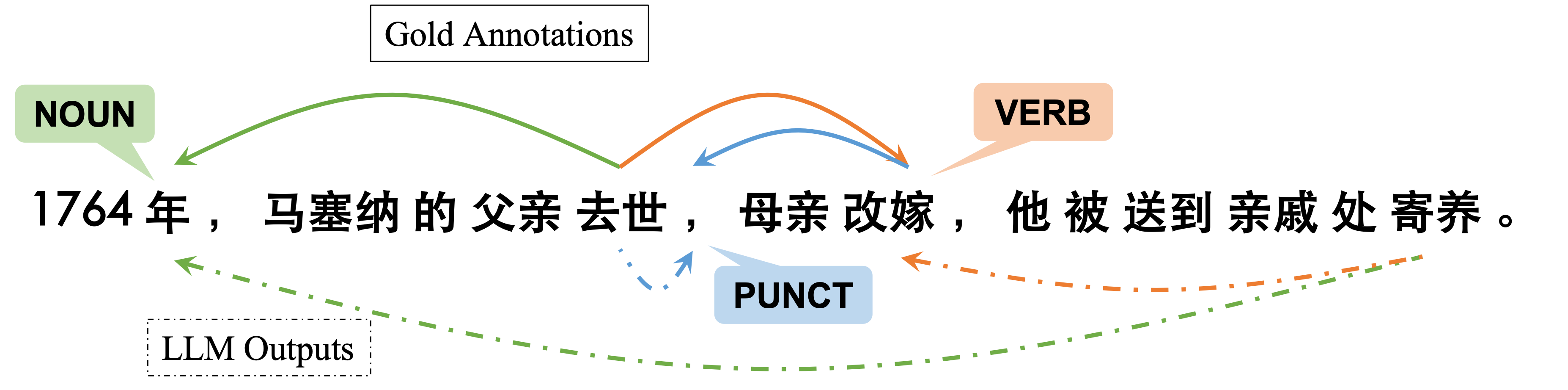}
    \caption{An example illustrating common errors for Chinese dependency parsing.}
    \label{fig:zh-error}
\end{figure}

Figure~\ref{fig:zh-error} illustrates an example that includes noun, verb, and punctuation errors, highlighting the challenge of analyzing paratactic structures with minimal syntactic markers.

\if0
This phenomenon highlights a broader challenge in parsing Chinese: the prevalence of paratactic structures with limited overt markers of syntactic hierarchy.  
Unlike many Indo-European languages, Chinese often lacks explicit function words to indicate syntactic roles, which leads to multiple plausible interpretations even in gold-standard annotations, making it difficult for LLMs to converge on a single correct structure.
\fi

\paragraph{Performance in other tasks.}

An LLM fine-tuned for dependency parsing clearly performs worse on other tasks, even if the base model has been instruction-tuned.
This performance degradation in general tasks may be mitigated or even reversed by fine-tuning the model on the dependency parsing task simultaneously with other instruction-tuning datasets \citep{asada-2025}; however, experimental verification remains a future challenge.

\subsection{Unimplemented UD tasks}

\paragraph{Tokenization.}

In the early stages of this work, we evaluated LLM-based word segmentation by inserting a word segmentation step at the beginning of step-by-step instructions.
However, particularly for Japanese, the segmentation accuracy was significantly lower than that of commonly used morphological analyzers.
To address this issue, full-parameter LLM training, including the word embedding layer, on large-scale training data would be necessary.
However, the associated cost could be several orders of magnitude higher than that of LoRA-SFT, which is employed in this study.
Thus, an efficient method for training word segmentation criteria tailored to LLMs is still required.

\paragraph{Lemmatization.}

Lemmatization has traditionally relied on dictionaries and heuristic rules; however, end-to-end approaches have recently gained traction \citep{qi-2020}.
LLMs may also be capable of effectively selecting the appropriate normalized form from a range of synonymous expressions or character variants by leveraging the knowledge acquired through large-scale pre-training, although this remains to be empirically validated.

\paragraph{Morphological features.}

The Universal Features\footnote{\href{https://universaldependencies.org/u/feat/}{https://universaldependencies.org/u/feat/}}
inventories over 200 lexical and inflectional features designed to classify word properties.
Decoder-based classifiers offer significant advantages for simultaneously classifying this large number of features, whereas using generative models such as LLMs is relatively inefficient.

\section{Conclusions}

We proposed a novel step-by-step prompting strategy for LLM-based dependency parsing using a simple tabular format, achieving improved output validity and parsing accuracy across 17 languages.
Multilingual SFT often outperformed monolingual models and generalized well to unseen languages.

\section*{Acknowledgments}

This work  was conducted as part of a collaborative research project between Recruit Co., Ltd. and the National Institute for Japanese Language and Linguistics.
We are grateful to all those involved in the management and support of this project.
We would also like to express our sincere gratitude to Yuji Matsumoto of RIKEN AIP for his valuable advice from the early stages of this research.
Finally, we thank the anonymous reviewers for their constructive and detailed comments.

\bibliography{custom}

\appendix

\section{Dataset Statistics}
\label{app:dataset-statistics}

Statistics for the Universal Dependencies treebanks used in the experiments are shown in Table \ref{tab:dataset-statistics}.

\begin{table*}[h!]
\centering
\small
\begin{tabular}{ll|rr|rr|rr}
\hline
\multirow{3}{*}{Language} & \multirow{3}{*}{Dataset} & \multicolumn{2}{c|}{Train} & \multicolumn{2}{c|}{Dev} & \multicolumn{2}{c}{Test} \\
& & number of & number of & number of & number of & number of & number of \\
& & sentences & tokens & sentences & tokens & sentences & tokens \\
\hline
UD r2.2: & & & & & & & \\
English & en\_ewt & 12,543 & 204,585 & 2,002 & 25,148 & 2,077 & 25,096 \\
\hline
UD r2.15: & & & & & & & \\
Arabic & ar\_padt & 6,075 & 223,881 & 909 & 30,239 & 680 & 28,264 \\
Bulgarian & bg\_btb & 8,907 & 124,336 & 1,115 & 16,089 & 1,116 & 15,724 \\
Catalan & ca\_ancora & 13,123 & 429,578 & 1,709 & 58,073 & 1,846 & 59,610 \\
Czech & cs\_pdt & 68,491 & 1,173,285 & 9,270 & 159,283 & 10,146 & 173,918 \\
German & de\_gsd & 13,814 & 263,791 & 799 & 12,480 & 977 & 16,498 \\
English & en\_ewt & 12,544 & 204,579 & 2,001 & 25,149 & 2,077 & 25,094 \\
Spanish & es\_ancora & 14,287 & 453,039 & 1,654 & 53,476 & 1,721 & 53,622 \\
French & fr\_gsd & 14,450 & 354,652 & 1,476 & 35,721 & 416 & 10,018 \\
Italian & it\_isdt & 13,121 & 276,014 & 564 & 11,907 & 482 & 10,417 \\
Japanese & ja\_gsd & 7,050 & 168,333 & 507 & 12,287 & 543 & 13,034 \\
Korean & ko\_gsd & 4,400 & 56,687 & 950 & 11,958 & 989 & 11,677 \\
Dutch & nl\_alpino & 12,289 & 186,027 & 718 & 11,541 & 596 & 11,046 \\
Norwegian & no\_bokmaal & 15,696 & 243,886 & 2,409 & 36,369 & 1,939 & 29,966 \\
Romanian & ro\_rrt & 8,043 & 185,125 & 752 & 17,073 & 729 & 16,324 \\
Russian & ru\_syntagrus & 69,630 & 1,204,640 & 8,906 & 153,325 & 8,800 & 157,718 \\
Slovenian & sl\_ssj & 10,903 & 215,155 & 1,250 & 26,500 & 1,282 & 25,442 \\
Simplified Chinese & zh\_gsdsimp & 3,997 & 98,616 & 500 & 12,663 & 500 & 12,012 \\
\hline
\end{tabular}
\caption{Statistics of Universal Dependencies treebanks used in SFT experiments.}
\label{tab:dataset-statistics}
\end{table*}

\section{Contamination Verification}
\label{app:contamination-verification}

A major concern in LLM-based evaluation is the contamination of testing data \citep{shokri-2017,das-2025}.  
To address this, we employed two diagnostics: (1) observing learning curves on \texttt{UD\_English-EWT r2.15} to detect unusually high initial performance, and (2) comparing fine-tuning results using training-only vs. training + test data.  
Evaluation results for contamination verification are presented below.

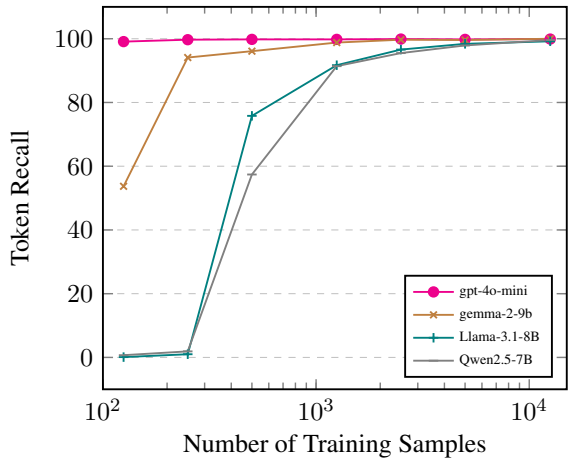
\begin{figure}[t!]
    \centering
    \resizebox{\linewidth}{!}{
    \tikzset{every mark/.append style={scale=1.0}}
    \begin{tikzpicture}
    \begin{axis}[
        xmode=log,
        xlabel={Number of Training Samples},
        ylabel={Token Recall},
        xmin=100, xmax=15000,
        ymin=-10, ymax=110,
        ytick={0,20,40,60,80,100},
        legend pos=south east,
        legend style={font=\tiny},
        legend cell align={left},
        ymajorgrids=true,
        grid style=dashed,
    ]

    \pgfplotsset{every axis/.append style={
    thick,
    tick style={semithick}}}

    \addplot[
        color=magenta,
        mark=*,
        ]
        coordinates {
        (125,	99.1)
        (250,	99.7)
        (500,	99.8)
        (1250,	99.8)
        (2500,	99.9)
        (5000,	99.8)
        (12544,	99.9)
        };
    \addplot[
        color=brown,
        mark=x,
        ]
        coordinates {
        (125,	53.7)
        (250,	94.1)
        (500,	96.1)
        (1250,	98.8)
        (2500,	99.7)
        (5000,	99.6)
        (12544,	100.0)
        };
    \addplot[
        color=teal,
        mark=+,
        ]
        coordinates {
        (125,	0.1)
        (250,	1.0)
        (500,	75.8)
        (1250,	91.7)
        (2500,	96.6)
        (5000,	98.4)
        (12544,	99.2)
        };
    \addplot[
        color=gray,
        mark=-,
        ]
        coordinates {
        (125, 0.7)
        (250, 1.9)
        (500, 57.4)
        (1250, 91.4)
        (2500, 95.5)
        (5000, 97.9)
        (12544, 99.6)
        };
        \legend{gpt-4o-mini, gemma-2-9b, Llama-3.1-8B, Qwen2.5-7B}
        
    \end{axis}
    \end{tikzpicture}
    }
    \caption{Learning curve - token recall.}
    \label{fig:learning-curve-token}
\end{figure}

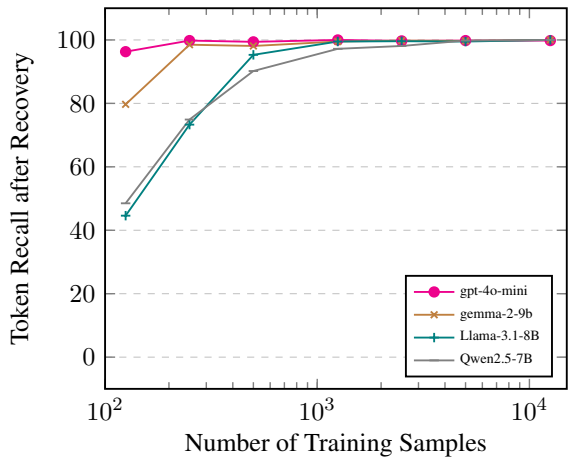
\begin{figure}[t!]
    \centering
    \resizebox{\linewidth}{!}{
    \tikzset{every mark/.append style={scale=1.0}}
    \begin{tikzpicture}
    \begin{axis}[
        xmode=log,
        xlabel={Number of Training Samples},
        ylabel={Token Recall after Recovery},
        xmin=100, xmax=15000,
        ymin=-10, ymax=110,
        ytick={0,20,40,60,80,100},
        legend pos=south east,
        legend style={font=\tiny},
        legend cell align={left},
        ymajorgrids=true,
        grid style=dashed,
    ]

    \pgfplotsset{every axis/.append style={
    thick,
    tick style={semithick}}}

    \addplot[
        color=magenta,
        mark=*,
        ]
        coordinates {
        (125, 96.3)
        (250, 99.8)
        (500, 99.4)
        (1250, 100.0)
        (2500, 99.7)
        (5000, 99.8)
        (12544, 99.8)
        };
    \addplot[
        color=brown,
        mark=x,
        ]
        coordinates {
        (125, 79.7)
        (250, 98.5)
        (500, 98.1)
        (1250, 99.5)
        (2500, 99.8)
        (5000, 99.7)
        (12544, 100.0)
        };
    \addplot[
        color=teal,
        mark=+,
        ]
        coordinates {
        (125, 44.6)
        (250, 73.3)
        (500, 95.3)
        (1250, 99.5)
        (2500, 99.6)
        (5000, 99.5)
        (12544, 100.0)
        };
    \addplot[
        color=gray,
        mark=-,
        ]
        coordinates {
        (125, 48.5)
        (250, 74.9)
        (500, 90.2)
        (1250, 97.2)
        (2500, 98.1)
        (5000, 99.8)
        (12544, 100.0)        };
        \legend{gpt-4o-mini, gemma-2-9b, Llama-3.1-8B, Qwen2.5-7B}
        
    \end{axis}
    \end{tikzpicture}
    }
    \caption{Learning curve - token recall after recovery.}
    \label{fig:learning-curve-token-after-recovery}
\end{figure}

\paragraph{Learning curves.}
Prior to the analysis, the learning curves of token recall (Figure~\ref{fig:learning-curve-token}) show that  \texttt{gpt-4o-mini} is able to generate outputs with correct formats in very early stage, while other models need to be trained, and the learning curves of token recall after recovery (Figure~\ref{fig:learning-curve-token-after-recovery}) indicates our simple recovery algorithm works effectively. 

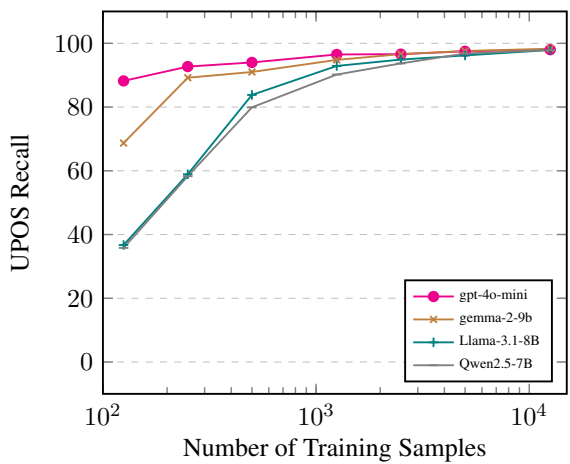
\begin{figure}[b!]
    \centering
    \resizebox{\linewidth}{!}{
    \tikzset{every mark/.append style={scale=1.0}}
    \begin{tikzpicture}
    \begin{axis}[
        xmode=log,
        xlabel={Number of Training Samples},
        ylabel={UPOS Recall},
        xmin=100, xmax=15000,
        ymin=-10, ymax=110,
        ytick={0,20,40,60,80,100},
        legend pos=south east,
        legend style={font=\tiny},
        legend cell align={left},
        ymajorgrids=true,
        grid style=dashed,
    ]

    \pgfplotsset{every axis/.append style={
    thick,
    tick style={semithick}}}

    \addplot[
        color=magenta,
        mark=*,
        ]
        coordinates {
        (125, 88.2)
        (250, 92.7)
        (500, 94.0)
        (1250, 96.5)
        (2500, 96.6)
        (5000, 97.5)
        (12544, 98.0)        };
    \addplot[
        color=brown,
        mark=x,
        ]
        coordinates {
        (125, 68.7)
        (250, 89.2)
        (500, 91.0)
        (1250, 94.8)
        (2500, 96.6)
        (5000, 97.6)
        (12544, 98.3)
        };
    \addplot[
        color=teal,
        mark=+,
        ]
        coordinates {
        (125, 36.7)
        (250, 58.9)
        (500, 83.8)
        (1250, 92.9)
        (2500, 94.9)
        (5000, 96.2)
        (12544, 97.9)
        };
    \addplot[
        color=gray,
        mark=-,
        ]
        coordinates {
        (125, 35.8)
        (250, 58.3)
        (500, 79.9)
        (1250, 90.2)
        (2500, 93.7)
        (5000, 96.8)
        (12544, 97.9)
        };
        \legend{gpt-4o-mini, gemma-2-9b, Llama-3.1-8B, Qwen2.5-7B}
        
    \end{axis}
    \end{tikzpicture}
    }
    \caption{Learning curve - UPOS recall.}
    \label{fig:learning-curve-upos-recall}
\end{figure}

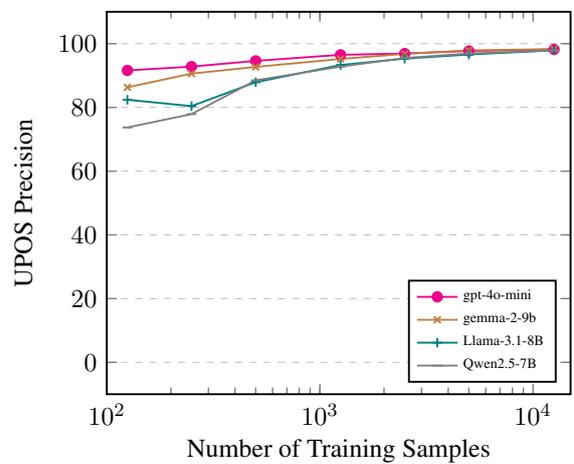
\begin{figure}[b!]
    \centering
    \resizebox{\linewidth}{!}{
    \tikzset{every mark/.append style={scale=1.0}}
    \begin{tikzpicture}
    \begin{axis}[
        xmode=log,
        xlabel={Number of Training Samples},
        ylabel={UPOS Precision},
        xmin=100, xmax=15000,
        ymin=-10, ymax=110,
        ytick={0,20,40,60,80,100},
        legend pos=south east,
        legend style={font=\tiny},
        legend cell align={left},
        ymajorgrids=true,
        grid style=dashed,
    ]

    \pgfplotsset{every axis/.append style={
    thick,
    tick style={semithick}}}

    \addplot[
        color=magenta,
        mark=*,
        ]
        coordinates {
        (125, 91.6)
        (250, 92.8)
        (500, 94.6)
        (1250, 96.5)
        (2500, 96.9)
        (5000, 97.7)
        (12544, 98.2)
        };
    \addplot[
        color=brown,
        mark=x,
        ]
        coordinates {
        (125, 86.3)
        (250, 90.6)
        (500, 92.7)
        (1250, 95.2)
        (2500, 96.8)
        (5000, 97.9)
        (12544, 98.3)        };
    \addplot[
        color=teal,
        mark=+,
        ]
        coordinates {
        (125, 82.4)
        (250, 80.4)
        (500, 87.9)
        (1250, 93.3)
        (2500, 95.3)
        (5000, 96.6)
        (12544, 97.9)
        };
    \addplot[
        color=gray,
        mark=-,
        ]
        coordinates {
        (125, 73.7)
        (250, 77.9)
        (500, 88.5)
        (1250, 92.8)
        (2500, 95.5)
        (5000, 96.9)
        (12544, 97.9)
        };
        \legend{gpt-4o-mini, gemma-2-9b, Llama-3.1-8B, Qwen2.5-7B}
        
    \end{axis}
    \end{tikzpicture}
    }
    \caption{Learning curve - UPOS precision.}
    \label{fig:learning-curve-upos-precision}
\end{figure}

For the learning curves of UPOS recall (Figure~\ref{fig:learning-curve-upos-recall}), the similarity between Figure~\ref{fig:learning-curve-token-after-recovery} and Figure~\ref{fig:learning-curve-upos-recall} suggests that the UPOS tagging task is one of the abilities that has been acquired in advance in these LLMs, which is also indicated by the high initial accuracy of the precision-based learning curves of UPOS in Figure~\ref{fig:learning-curve-upos-precision}.

In contrast, the gradual learning curves for HEAD and DEPREL identification (Figures~\ref{fig:learning-curve-head-recall} and~\ref{fig:learning-curve-head-deprel-recall}) indicate the necessity of SFT for learning the knowledge for dependency parsing.

Overall, we conclude that the tested LLMs do not exhibit potential contamination in syntactic head and relational label identification tasks, despite possible prior exposure to the UPOS tagging task.

\paragraph{Effect of additional training on test set.}
In Table~\ref{tab:upos-las-diff}, when the testing data was included in training, all models naturally achieved certain accuracy gains (>+0.7 for UPOS, >+2.3 for LAS).
This indicates a low possibility of contamination for the  test set of \texttt{UD\_English-EWT r2.15}.

\begin{table}[h!]
\centering
\resizebox{\linewidth}{!}{
\begin{tabular}{l|cc|cc}
\hline
  & \multicolumn{2}{c|}{UPOS} & \multicolumn{2}{c}{LAS} \\
\hline
  & \multirow{2}{*}{train} & train + test & \multirow{2}{*}{train} & train + test  \\
  & & ($\pm$ diff) & & ($\pm$ diff)\\
\hline
\texttt{gpt-4o-mini-2024-07-18} & 98.3 & 99.3 (+1.0) & 93.2 & 97.1 (+3.9) \\
\texttt{Qwen2.5-7B}             & 97.1 & 98.5 (+1.4) & 88.7 & 93.6 (+4.9) \\
\texttt{Llama-3.1-8B}           & 97.2 & 98.7 (+1.5) & 88.9 & 94.1 (+5.2) \\
\texttt{gemma-2-9b}             & 98.0 & 98.7 (+0.7) & 91.7 & 94.0 (+2.3) \\
\hline
\end{tabular}
}
\caption{Difference of UPOS and LAS scores between SFT on training data only and training + test data for \texttt{UD\_English-EWT r2.15}.}
\label{tab:upos-las-diff}
\end{table}

\clearpage

\begin{figure}[b!]
    \centering
    \resizebox{\linewidth}{!}{
    \tikzset{every mark/.append style={scale=1.0}}
    \begin{tikzpicture}
    \begin{axis}[
        xmode=log,
        xlabel={Number of Training Samples},
        ylabel={HEAD Recall},
        xmin=100, xmax=15000,
        ymin=-10, ymax=110,
        ytick={0,20,40,60,80,100},
        legend pos=south east,
        legend style={font=\tiny},
        legend cell align={left},
        ymajorgrids=true,
        grid style=dashed,
    ]

    \pgfplotsset{every axis/.append style={
    thick,
    tick style={semithick}}}

    \addplot[
        color=magenta,
        mark=*,
        ]
        coordinates {
        (125, 72.6)
        (250, 81.7)
        (500, 86.4)
        (1250, 89.4)
        (2500, 91.2)
        (5000, 92.9)
        (12544, 94.1)
        };
    \addplot[
        color=brown,
        mark=x,
        ]
        coordinates {
        (125, 48.8)
        (250, 68.8)
        (500, 76.4)
        (1250, 87.4)
        (2500, 90.4)
        (5000, 92.7)
        (12544, 94.8)
        };
    \addplot[
        color=teal,
        mark=+,
        ]
        coordinates {
        (125, 8.0)
        (250, 41.1)
        (500, 62.9)
        (1250, 80.2)
        (2500, 86.2)
        (5000, 89.7)
        (12544, 93.6)
        };
    \addplot[
        color=gray,
        mark=-,
        ]
        coordinates {
        (125, 19.2)
        (250, 38.9)
        (500, 56.0)
        (1250, 75.8)
        (2500, 83.3)
        (5000, 89.4)
        (12544, 93.3)
        };
        \legend{gpt-4o-mini, gemma-2-9b, Llama-3.1-8B, Qwen2.5-7B}
        
    \end{axis}
    \end{tikzpicture}
    }
    \caption{Learning curve - HEAD recall.}
    \label{fig:learning-curve-head-recall}
\end{figure}

\begin{figure}[b!]
    \centering
    \resizebox{\linewidth}{!}{
    \tikzset{every mark/.append style={scale=1.0}}
    \begin{tikzpicture}
    \begin{axis}[
        xmode=log,
        xlabel={Number of Training Samples},
        ylabel={HEAD + DEPREL Recall},
        xmin=100, xmax=15000,
        ymin=-10, ymax=110,
        ytick={0,20,40,60,80,100},
        legend pos=south east,
        legend style={font=\tiny},
        legend cell align={left},
        ymajorgrids=true,
        grid style=dashed,
    ]

    \pgfplotsset{every axis/.append style={
    thick,
    tick style={semithick}}}

    \addplot[
        color=magenta,
        mark=*,
        ]
        coordinates {
        (125, 66.1)
        (250, 76.2)
        (500, 81.8)
        (1250, 86.0)
        (2500, 88.7)
        (5000, 90.6)
        (12544, 92.3)
                };
    \addplot[
        color=brown,
        mark=x,
        ]
        coordinates {
        (125, 41.5)
        (250, 61.2)
        (500, 70.4)
        (1250, 83.4)
        (2500, 87.6)
        (5000, 90.4)
        (12544, 93.3)
                };
    \addplot[
        color=teal,
        mark=+,
        ]
        coordinates {
        (125, 5.2)
        (250, 33.5)
        (500, 55.3)
        (1250, 74.2)
        (2500, 82.2)
        (5000, 86.9)
        (12544, 91.6)
                };
    \addplot[
        color=gray,
        mark=-,
        ]
        coordinates {
        (125, 11.8)
        (250, 28.9)
        (500, 47.4)
        (1250, 69.4)
        (2500, 79.1)
        (5000, 86.7)
        (12544, 91.4)
                };
        \legend{gpt-4o-mini, gemma-2-9b, Llama-3.1-8B, Qwen2.5-7B}
        
    \end{axis}
    \end{tikzpicture}
    }
    \caption{Learning curve - HEAD+DEPREL recall.}
    \label{fig:learning-curve-head-deprel-recall}
\end{figure}

\clearpage

\section{Prompt Examples}
\label{app:prompt_examples}

The examples of Chain-of-Thought instruction prompts used in the experiments are shown in Figure \ref{fig:single-step-prompt} and Figure \ref{fig:three-step-prompt}.

\begin{figure}[h!]
\begin{center}
\footnotesize
\begin{verbatim}
[system]
You are an English linguist and specialize in
English dependency analysis based on Universal
Dependencies.

[user]
We will now perform dependency parsing on English
sentence. After splitting the input sentence into
words as shown below, create a TSV with five fields:
word index from 1 to 6 + word +  the Universal
Dependencies part of speech + the dependent word
index + the Universal Dependencies relation.
However, for the word that is the main predicate of
the sentence, the dependent word index should be 0.

input sentence:
I'm just speculating now.

words:
I 
'm 
just 
speculating 
now
. 

[assistant]
1	I 	PRON	4	nsubj
2	'm 	AUX	4	aux
3	just 	ADV	4	advmod
4	speculating 	VERB	0	root
5	now	ADV	4	advmod
6	. 	PUNCT	4	punct
\end{verbatim}
\caption{Example for 1-step parsing instruction.}
\label{fig:single-step-prompt}
\end{center}
\end{figure}

\begin{figure}[h!]
\begin{center}
\footnotesize
\begin{verbatim}
[system]
You are an English linguist and specialize in
English dependency analysis based on Universal
Dependencies.

[user]
We will now perform dependency parsing on English
sentence. After splitting the input sentence into
words as shown below, execute following three
tasks:

- Task 1
Create a TSV with three fields: word index from 1
to 6 + word + part of speech.

- Task 2
Add a field for the dependent word indexes to
each row to the output of Task 1. However, for
the word that is the main predicate of the
sentence, the dependent word index should be 0.

- Task 3
Add a field for the Universal Dependencies
relation labels to the output of Task 2.

input sentence:
I'm just speculating now.

words:
I 
'm 
just 
speculating 
now
. 

[assistant]
- Task 1
1	I 	PRON
2	'm 	AUX
3	just 	ADV
4	speculating 	VERB
5	now	ADV
6	. 	PUNCT

- Task 2
1	I 	PRON	4
2	'm 	AUX	4
3	just 	ADV	4
4	speculating 	VERB	0
5	now	ADV	4
6	. 	PUNCT	4

- Task 3
1	I 	PRON	4	nsubj
2	'm 	AUX	4	aux
3	just 	ADV	4	advmod
4	speculating 	VERB	0	root
5	now	ADV	4	advmod
6	. 	PUNCT	4	punct
\end{verbatim}
\caption{Example for 3-step parsing instruction.}
\label{fig:three-step-prompt}
\end{center}
\end{figure}

\end{document}